\documentclass[11pt]{article}
\usepackage[numbers,sort&compress]{natbib}
\usepackage{amsmath,amssymb,graphicx,booktabs,xcolor,lineno}
\usepackage[hidelinks]{hyperref}
\usepackage{geometry}
\newcommand{\unit}{mmHg$\cdot$min}
\newcommand{\TrainWindows}{70,402}
\newcommand{\ValWindows}{14,270}
\newcommand{\TestWindows}{15,575}
\newcommand{\TotalWindows}{100,247}
\newcommand{\UsableCases}{3,423}
\newcommand{\UsableSubjects}{3,304}
\newcommand{\NoWindowCases}{41}
\newcommand{\ResidualLoss}{2.256}
\newcommand{\PersistenceLoss}{2.900}
\newcommand{\DirectLoss}{2.196}
\newcommand{\DetailLoss}{2.193}
\newcommand{\PrimaryDifference}{-0.644}
\newcommand{\PrimaryLower}{-0.814}
\newcommand{\PrimaryUpper}{-0.487}

\begin{document}
\title{Forecasting cumulative intraoperative hypotension burden: patient-disjoint evaluation of quantile boosting and alert policies}
\author{Yunfan Zhao\\GE Healthcare}
\date{}
\maketitle
\begin{abstract}
\textbf{Background:} Binary hypotension forecasts do not describe the depth and duration of future low pressure. Burden forecasts may support graded early warning, but better forecast scores need not yield more useful alerts.
\textbf{Objective:} To evaluate conditional quantiles of the next 15-minute mean arterial pressure (MAP) deficit below 65~mmHg and distinguish forecasting performance from episode-warning performance.
\textbf{Methods:} Public VitalDB numeric arterial MAP was reduced to 10-second medians without forward filling. Subject-disjoint training, validation and test splits included \TrainWindows, \ValWindows\ and \TestWindows\ labeled windows. We compared climatology, calibrated persistence, last-MAP-only boosting, and direct and residual quantile boosting, with and without additional MAP-history features. Patient-weighted pinball loss and subject-bootstrap intervals evaluated forecasting. Validation selected alert thresholds before a hash-recorded test evaluation.
\textbf{Results:} In the test split, the prespecified residual-versus-persistence comparison at the 90th percentile yielded losses of \ResidualLoss\ and \PersistenceLoss\ \unit{}, respectively (difference \PrimaryDifference; 95\% interval \PrimaryLower\ to \PrimaryUpper). Direct prediction achieved \DirectLoss, and direct prediction with additional MAP features achieved \DetailLoss. Episode-warning results were sensitive to the alert rule and operating threshold.
\textbf{Conclusions:} This retrospective benchmark separates burden-forecast skill from warning-policy performance. Findings support further study of direct burden estimation; they do not establish patient benefit, prospective reliability or transportability beyond the source hospital.
\end{abstract}

\section*{Summary table}
\textbf{What was already known}
\begin{itemize}
\item Depth and duration of intraoperative hypotension are relevant to clinical risk, and current pressure is a necessary forecasting benchmark.
\item Prediction accuracy and the operational behaviour of an alert policy are distinct evaluation targets.
\end{itemize}
\textbf{What this study adds}
\begin{itemize}
\item A reproducible benchmark for conditional quantiles of future cumulative pressure deficit, with patient-disjoint evaluation.
\item Matched direct and residual models, a fine-resolution history ablation, and a weight-scale sensitivity analysis.
\item Activation-based alert evaluation that prevents a continuously active alert from claiming multiple episodes.
\end{itemize}

\section{Introduction}
Biomedical research spans infectious disease \cite{xie2022escape}. At the bedside, a different challenge is to anticipate changes in routinely monitored physiology.

Intraoperative hypotension is an important monitoring problem. Resource-constrained maternal monitoring provides a related setting in which vital-sign acquisition and allocation must be considered together \cite{boehmer2025monitoring}. Observational studies link low mean arterial pressure (MAP) to postoperative injury, with associations depending on the pressure level and exposure duration \cite{salmasi2017,walsh2013}. A binary prediction of whether pressure will cross a threshold cannot distinguish a brief shallow dip from a prolonged deep excursion. Forecasting cumulative deficit provides a complementary representation of future exposure. It remains a physiological benchmark, rather than a validated surrogate for patient harm. Associations with postoperative acute kidney injury support considering sustained low-pressure exposure \cite{sun2015}.

Reproductive biology is another domain of biomedical investigation \cite{liu2026tmem95}. Physiological forecasting likewise requires evidence appropriate to its intended use, although its endpoints and validation procedures are distinct.

The broader clinical literature supports attention to both the depth and duration of low pressure. Mortality analyses also associate the severity and duration of intraoperative pressure abnormalities with postoperative risk \cite{monk2015}. A systematic review synthesised associations between intraoperative hypotension and postoperative adverse outcomes \cite{wesselink2018}; the Perioperative Quality Initiative addressed intraoperative pressure, risk and outcomes in elective surgery \cite{sessler2019}. These sources motivate measuring exposure, without establishing one threshold as appropriate for every patient. Updated consensus recommendations likewise place arterial-pressure management in the clinical context of the individual patient \cite{saugel2024}. The INPRESS trial evaluated individualised versus standard blood-pressure management in high-risk surgical patients \cite{futier2017}. Its intervention question differs from our retrospective forecasting question: predicting a pressure deficit is not equivalent to demonstrating benefit from treating it. A trial of higher intraoperative MAP targets did not demonstrate reduced adverse cardiovascular events \cite{wanner2021}. Separately, the POISE-3 comparison of hypotension-avoidance and hypertension-avoidance strategies found no difference in its primary vascular outcome \cite{marcucci2023}.

Prior prediction approaches include high-fidelity arterial-waveform machine learning \cite{hatib2018}, deep-learning models for intraoperative hypotension \cite{lee2021bja}, and the STEP-OP hybrid approach to five-minute prediction \cite{choe2021}. These studies establish relevant event-forecasting comparators, but their inputs, horizons and endpoints differ from the present numeric-MAP burden task. The HYPE randomised trial evaluated a warning system embedded in a clinical management workflow \cite{wijnberge2020}. That design illustrates the additional evidence needed to move from retrospective prediction to an intervention claim.

Immunological research further illustrates this breadth \cite{cao2022epitope}. The present study focuses instead on characterising future pressure exposure from clinical time series.

Medical informatics evaluation must connect such predictions to their use in decision support. Forecasting individual engagement in maternal-health programmes provides a complementary example of prediction intended to inform intervention allocation \cite{seow2024engagement}. A model may reduce average forecast loss while producing unstable or poorly timed alerts. Evaluation also depends on the sampling design: selecting cases and controls around pressure thresholds can make forecasting appear easier than scoring a consecutive series of eligible decision times. Prior work has raised concerns about selection bias in hypotension prediction \cite{enevoldsen2022}. Here, current-pressure persistence and a learned last-pressure-only model are explicit controls. Linear extrapolation of MAP has been studied as a simple comparator \cite{jacquet2022}, and a prospective observational comparison found similar predictive performance for the Hypotension Prediction Index and current-MAP thresholds in its setting \cite{mulder2024}. Studies comparing forward cohort and backward case-control validation \cite{davies2024}, examining selection bias in real-world samples \cite{yang2025bja}, and evaluating methods across centres \cite{chaari2026} show why the evaluation population must be specified alongside model performance. We therefore avoid importing published accuracy estimates as directly comparable benchmarks.

The conceptual analogy to downside-risk forecasting is limited and precise. We forecast an upper quantile of a nonnegative accumulated deficit, score it using an asymmetric loss, and backtest a decision rule against observed events. No monetary interpretation, financial saving or expected-shortfall estimate is implied. Quantile regression and gradient boosting are established methods \cite{koenker1978,chen2016xgboost}. Our contribution is their evaluation for an explicitly defined burden endpoint and the separation of forecast performance from event-warning performance. The distinction between estimation and downstream optimisation also motivates evaluating these stages separately \cite{elmachtoub2023estimate}.

This study evaluates supervised machine learning for a continuous physiological forecasting target. Its learned predictors are gradient-boosted trees trained on structured MAP-history features, rather than large language models (LLMs). The distinction matters for interpreting the contribution: the question is whether learning conditional burden quantiles improves forecasting and warning performance, not whether language generation can support bedside decisions.

We asked three questions: whether learned burden quantiles improve on calibrated persistence; whether residual learning or finer-resolution MAP history adds value under matched settings; and whether the resulting scores support useful episode warnings under fixed validation-selected policies. We used an independent test split within one hospital and did not perform external or prospective validation.

\section{Methods}
\subsection{Data source and reconstruction}
VitalDB provides de-identified intraoperative recordings and clinical metadata \cite{lee2022vitaldb,lee2022physionet}. We downloaded clinical metadata from PhysioNet version 1.0.0 and numeric tracks from the official VitalDB API. Vital Recorder describes the source infrastructure for time-synchronised perioperative signal acquisition \cite{lee2018recorder}; PhysioNet provides a broader framework for sharing physiological recordings and analysis resources \cite{goldberger2000}. The source reports institutional approval H-1408-101-605 and a waiver of consent for its collection and release. The present study is a noncommercial secondary analysis of these publicly available, de-identified data, with no participant recruitment or intervention.

The source contains 6,388 cases. We selected adults aged at least 18 years, general anaesthesia, a listed \texttt{Solar8000/ART\_MBP} channel, and nominal case duration of at least 60 minutes. Ages recorded as greater than 89 were represented as 90; the original flag was retained in metadata. We attempted all 3,626 candidate MAP tracks and cached downloads. The current reconstruction had no download failures, 161 cases with no screened values in the analysis interval and one invalid interval. Of 3,464 processed cases, \UsableCases\ contributed at least one labeled forecasting window. The remaining \NoWindowCases\ did not meet the window criteria. Explicit cohort derivation and eligibility reporting are particularly important for analyses of routinely collected health data \cite{benchimol2015}.

This study continues an earlier handoff whose underlying code, predictions and caches were unavailable. Its reported validation scores are treated as historical development evidence, not as outputs reproduced by the current pipeline. The reconstruction uses a larger successfully downloaded cohort, XGBoost 2.1.4, explicit training-weight normalisation and a revised episode definition. Results from the two implementations are not numerically interchangeable.

\subsection{Signal processing and burden target}
We restricted samples to $[\max(0,\mathrm{anestart}),\min(\mathrm{caseend},\mathrm{aneend}))$ and retained finite values from 20 to 200~mmHg. This is a plausibility screen, not validated artefact adjudication. Arterial-catheter measurement also depends on levelling, zeroing and the dynamic response of the monitoring system \cite{saugel2020}. The median of available samples in each 10-second bin was computed without forward filling. A complete minute required six observed bin medians. Its MAP was their median; its burden was
\begin{equation}
b_m=\frac{1}{6}\sum_{k=1}^{6}(65-x_{m,k})_+,\qquad (a)_+=\max(a,0).
\end{equation}
The forecast origin $t$ is a minute boundary. History comprises minutes $[t-30,t)$, and the target is
\begin{equation}
B_{15}(t)=\sum_{m=t}^{t+14}b_m,
\end{equation}
in \unit{}. All 15 future minutes must be complete for a labeled window. Computing deficit before minute aggregation retains excursions obscured by a minute median. The 65~mmHg threshold defines this benchmark and is not an individualised treatment recommendation. Reported hypotension incidence depends strongly on the operational definition, so pressure and duration criteria must accompany comparisons \cite{bijker2007}.

Origins occur every five minutes relative to the first minute bin of the anaesthesia interval. Using the anaesthesia interval includes pre-incision exposure, which has also been examined in relation to postoperative kidney injury \cite{maheshwari2018}. History eligibility requires at least 27 of the preceding 30 minutes to be complete and the latest minute to be present. History-eligible origins without complete future labels remain in alert replay. However, replay still uses retrospective anaesthesia boundaries and ends 15 minutes before the end of the available timeline; it is not a verified real-time availability simulation.

\subsection{Splitting, features and weighting}
The first eight hexadecimal digits of SHA-256 of the string \texttt{vitaldb-mie2027-v1:} followed by the integer subject identifier were divided by $2^{32}$. Values below 0.70, from 0.70 to below 0.85, and at least 0.85 assigned training, validation and test respectively. Repeated procedures therefore stay together. This deterministic split is not a calendar-time backtest. Separating subjects respects the hierarchical dependence structure of repeated observations; validation strategies for structured data require the split unit to match the intended generalisation question \cite{roberts2017}.

The base predictors are mean, standard deviation, minimum, maximum, linear slope and observed fraction over 5, 15 and 30 minutes, plus latest minute MAP, age and sex (21 features). The additional MAP-history set adds observed cumulative burden over 5, 15 and 30 minutes, minimum 10-second MAP and count of low 10-second bins over five minutes, and elapsed minutes from the first analysis bin (27 features total). Partially observed history burdens sum observed contributions; base coverage features remain available. No future observation enters these predictors. We did not collect additional drug or physiological channels for this experiment.

Each training subject receives equal total weight, rescaled so that the mean window weight is one. This scale matters because XGBoost's minimum child weight and regularisation are absolute quantities. Evaluation first averages each metric within subject and then across subjects. All reported coverage uses the same patient weighting as loss.

\subsection{Forecast models and development iteration}
For $q\in\{0.50,0.90,0.95\}$, the pinball loss is
\begin{equation}
\rho_q(y,\hat y)=\max\{q(y-\hat y),(q-1)(y-\hat y)\}.
\end{equation}
It scores the requested quantile \cite{koenker1978,gneiting2007}. At $q=0.90$, equal-sized underprediction receives nine times the penalty of overprediction. A quantile is not an expected-shortfall estimate.

Climatology uses subject-weighted training quantiles of burden. Persistence uses $P=15(65-\mathrm{MAP}_{\rm last})_+$ plus training quantiles of $B_{15}-P$. Learned models comprise last-MAP-only direct boosting; base-feature direct and residual boosting; and direct and residual boosting with additional MAP history. Residual models predict $B_{15}-P$ and add $P$ at inference. All outputs are clipped at zero. The residual formulation separates a fixed persistence component from a learned correction. Structurally constrained prediction adapters provide a methodological analogy for this separation \cite{wang2026embeddingfoundationmodelpredictions}; our model does not inherit their guarantees, and clipping enforces only nonnegative outputs.

Gradient boosting provides a general framework for additive function estimation \cite{friedman2001}. Quantile regression forests offer another established route to conditional quantiles \cite{meinshausen2006}; we cite them as methodological context, not as an evaluated comparator. XGBoost's native quantile objective used 200 rounds, depth 3, learning rate 0.04, minimum child weight 10, L2 penalty 10, row fraction 0.8, column fraction 0.9, histogram trees, two threads and seed 2027. Quantiles are separate heads with no sorting or calibration after fitting; their crossing frequency is audited. No model is refitted on validation before test evaluation. The pretrained-model extension in the inherited draft was not run and is not a contribution of this study. Scalable neural approximations provide additional architectural options \cite{zhao2024kernels}, but none was included in this tree-based benchmark.

The first validation comparison favoured direct prediction over residual prediction. We then ran a bounded validation-only sensitivity using unnormalised weights with total weight one per subject, keeping the same tree settings. This investigates a methodological source of ranking instability. It is exploratory, was performed before test evaluation, and did not replace the primary weight convention.

For a conditional mean pressure forecast, convexity implies
\begin{equation}
(65-\mathbb E[X\mid H])_+\leq\mathbb E[(65-X)_+\mid H].
\end{equation}
This motivates studying the burden functional directly. It does not establish superiority over a calibrated forecast-then-integrate model, which was not evaluated, and it does not apply as stated to an arbitrary median pressure forecast. Marginal pressure quantiles also do not specify the temporal dependence required for burden quantiles.

\subsection{Episodes and alert policies}
An episode begins with a run of at least two consecutive complete minute medians below 65~mmHg. Qualifying runs are then merged across at most one complete intervening minute. Isolated low minutes are not merged into qualifying runs. Episodes ending at the recording boundary or immediately before a missing minute are censored. A warning opportunity requires five complete normotensive preceding minutes, an uncensored episode, and at least one history-eligible decision origin 1--15 minutes before onset.

We replayed threshold, two-consecutive-origin persistence, and hysteresis rules on the 90th-percentile burden score. Hysteresis deactivates below half its activation threshold. A gap in the five-minute decision grid resets state; a terminal reset is counted as a deactivation. A new activation matches the earliest unmatched eligible episode 1--15 minutes ahead. Activations during an ongoing episode, duplicates, and unmatched activations are counted separately. A sustained alert receives no additional episode credit until it deactivates and activates again.

Sensitivity is captured eligible episodes divided by eligible episodes. Precision is undefined when no alert activates; tables show a dash in that case. Precision divides captures by all activations, including ongoing-episode and duplicate activations. Unmatched activations define the false-alert numerator. Exposure is five minutes per history-eligible origin, termed \emph{decision-exposure hours}; it is not independently established continuous monitoring time. This conditional denominator and exclusion of unavailable origins must accompany interpretation of false-alert rates.

Threshold candidates comprised 31 validation score quantiles plus an alert-off setting. For each policy and score, validation selected maximum sensitivity subject to 0.10, 0.25 or 0.50 unmatched alerts per decision-exposure hour; ties favoured fewer activations and then a higher threshold. These are budget-constrained comparisons, not exact matching of achieved total alert rates. Thresholds were frozen and applied unchanged to test. The full grid is retained so that a reduction in turnover can be examined alongside missed events. This replay scores alert timing against observed episodes; it does not estimate outcomes under alternative treatments, as distinct off-policy evaluation problems require \cite{elmachtoub2023balanced}.

\subsection{Statistical analysis and test freeze}
The primary contrast, inherited from the earlier study question, was base-feature residual boosting versus persistence at $q=0.90$ on all labeled test origins. Other models, quantiles, subgroups and alert comparisons are secondary or exploratory. Paired percentile intervals use 2,000 subject-bootstrap replicates with seed 2027. The bootstrap provides a resampling approach to uncertainty estimation \cite{efron1979}; in this study the resampling unit is the subject rather than an overlapping window. Subjects retain all their windows and cases. Retaining each sampled cluster intact reflects the dependence addressed by clustered-data bootstrap methods \cite{field2007}. Alert sensitivity and false-rate intervals also resample subjects. No multiplicity correction is applied to exploratory comparisons.

The onset subgroup requires five complete preceding minute medians at least 65~mmHg; it does not exclude all 10-second dips. A no-jump sensitivity removes windows with adjacent 10-second median changes above 30~mmHg anywhere in history or future. This future-based filter is an evaluation sensitivity, not a deployable predictor. Age, sex and pre-origin volatility strata are exploratory; the volatility threshold is the training median of 15-minute MAP standard deviation.

Before test predictions, the dataset, saved models, analysis scripts, event definitions, primary contrast and validation-selected thresholds were recorded with SHA-256 hashes. The evaluator verifies hashes and creates a one-use marker. Version-specific evaluation is also motivated by evidence of checkpoint-dependent benchmark scores in chat LLMs \cite{fan2026movingtargetlongitudinalaudit}; here, the freeze identifies the exact XGBoost models and analysis artefacts evaluated. All targets were independently recomputed from cached 10-second bins. We also checked unique origins, row alignment and subject disjointness, and tested alert behaviour across gaps and sustained activations.

\section{Results}
\subsection{Reconstructed cohort}
\begin{table}[tbp]
\centering\small
\caption{Reconstructed labeled-window cohort. Positive windows have burden greater than zero; overlapping windows are not independent events.}
\begin{tabular}{lrrrr}\toprule
Split & Cases & Subjects & Windows & Positive \\
\midrule
Train & 2,395 & 2,307 & 70,402 & 23,473 \\
Validation & 505 & 491 & 14,270 & 4,689 \\
Test & 523 & 506 & 15,575 & 4,924 \\
\bottomrule\end{tabular}
\end{table}

There were \TotalWindows\ labeled windows across \UsableCases\ cases and \UsableSubjects\ subjects. Among processed timelines, 721,231 of 804,424 candidate minutes were complete. These are signal-coverage counts, not adjudicated artefact-free minutes. The current cohort has no technical download exclusions. \begin{table}[tbp]
\centering\small
\caption{Operational episodes and eligible warning opportunities under the corrected run and censoring rules.}
\begin{tabular}{lrrr}\toprule
Split & Episodes & Opportunities & Subjects \\
\midrule
Train & 6798 & 2762 & 1135 \\
Validation & 1362 & 535 & 240 \\
Test & 1472 & 589 & 242 \\
\bottomrule\end{tabular}
\end{table}

\subsection{Forecasting and the development iteration}
\begin{table}[tbp]
\centering\small
\caption{Patient-weighted pinball loss (mmHg$\cdot$min). Validation informed development; test was evaluated after the analysis freeze.}
\begin{tabular}{lrrrr}\toprule
Model & Val. $q_{.90}$ & Val. $q_{.95}$ & Test $q_{.90}$ & Test $q_{.95}$ \\
\midrule
Climatology & 4.500 & 3.635 & 4.458 & 3.621 \\
Calibrated persistence & 2.910 & 2.240 & 2.900 & 2.262 \\
Last MAP only & 2.552 & 1.902 & 2.538 & 1.908 \\
Direct, base features & 2.239 & 1.590 & 2.196 & 1.596 \\
Residual, base features & 2.295 & 1.640 & 2.256 & 1.636 \\
Direct, additional MAP & 2.205 & 1.562 & 2.193 & 1.588 \\
Residual, additional MAP & 2.252 & 1.608 & 2.246 & 1.619 \\
\bottomrule\end{tabular}
\end{table}

The validation results prompted examination of the direct-versus-residual ranking rather than preservation of the inherited draft's preferred method. On validation, direct prediction with additional MAP history reduced 90th-percentile loss by 0.034 relative to the base direct model (difference -0.034; 95\% interval -0.065 to -0.007). With unnormalised subject weights, direct and residual 90th-percentile losses were 2.228 and 2.286; direct prediction remained better. Weight scaling alone therefore did not explain the changed ranking relative to the inherited handoff.

On the held-out test data, the primary residual-minus-persistence difference was \PrimaryDifference\ \unit{} (95\% interval \PrimaryLower\ to \PrimaryUpper). This estimates a difference in retrospective forecast loss, not clinical efficacy. Base direct prediction also outperformed residual prediction in a secondary paired comparison (difference $-0.060$; 95\% interval $-0.093$ to $-0.031$). The additional MAP features gave little further test improvement over base direct prediction at $q=0.90$ (difference $-0.003$; interval $-0.026$ to $0.019$), despite a favourable validation comparison. We therefore do not claim a confirmed benefit from that extension. \begin{table}[tbp]
\centering\small
\caption{Test sensitivities and patient-weighted coverage for the 90th-percentile head. The jump restriction is retrospective.}
\begin{tabular}{lrrrr}\toprule
Model & Onset loss & No-jump loss & $P(B< Q)$ \% & $P(B\leq Q)$ \% \\
\midrule
Calibrated persistence & 3.018 & 2.552 & 90.6 & 90.7 \\
Last MAP only & 2.595 & 2.249 & 90.9 & 90.9 \\
Direct, base features & 2.275 & 1.837 & 90.7 & 91.2 \\
Residual, base features & 2.267 & 1.923 & 90.8 & 91.2 \\
Direct, additional MAP & 2.252 & 1.811 & 90.8 & 91.3 \\
Residual, additional MAP & 2.255 & 1.897 & 90.9 & 91.3 \\
\bottomrule\end{tabular}
\end{table}
Across labeled test windows, any adjacent quantile crossing occurred in 0.28\% for base direct prediction and 0.14\% for direct prediction with additional MAP features. These are window-level diagnostic rates, not patient-weighted performance estimates.

\begin{figure}[tbp]\centering\includegraphics[width=\linewidth]{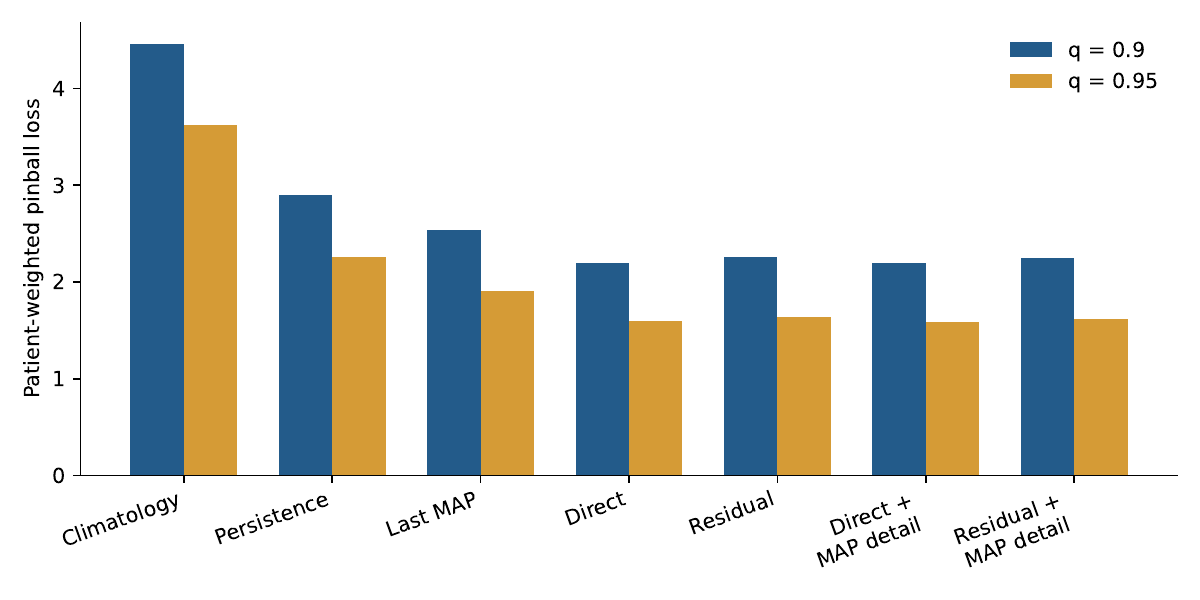}\caption{Patient-weighted test pinball losses. Lower is better. All models were fitted on training data only; full bootstrap contrasts are supplied with the analysis.}\end{figure}

\subsection{Alert evaluation}
\begin{table}[tbp]
\centering\small
\caption{Test alert results at the validation budget of 0.25 unmatched alerts per decision-exposure hour. False rates and total activation rates are distinct.}
\begin{tabular}{llrrrrr}\toprule
Score & Rule & Sens. \% & Prec. \% & False/h & Alerts/h & Turns/h \\
\midrule
Persistence score & Threshold & 1.9 & 1.6 & 0.03 & 0.46 & 0.91 \\
Persistence score & 2-origin & 0.0 & -- & 0.00 & 0.00 & 0.00 \\
Persistence score & Hysteresis & 1.9 & 1.6 & 0.03 & 0.44 & 0.89 \\
Direct base & Threshold & 13.4 & 9.7 & 0.16 & 0.53 & 1.06 \\
Direct base & 2-origin & 16.3 & 14.3 & 0.23 & 0.44 & 0.87 \\
Direct base & Hysteresis & 13.8 & 11.7 & 0.19 & 0.45 & 0.91 \\
Direct additional & Threshold & 10.9 & 8.2 & 0.15 & 0.51 & 1.02 \\
Direct additional & 2-origin & 16.3 & 14.0 & 0.23 & 0.45 & 0.89 \\
Direct additional & Hysteresis & 13.1 & 11.0 & 0.20 & 0.46 & 0.92 \\
\bottomrule\end{tabular}
\end{table}

The achieved test rate need not equal the validation budget. Policy comparisons therefore report actual false and total activation rates alongside sensitivity, precision and turnover. For direct prediction with additional MAP history, the two-origin rule captured 96 of 589 test opportunities: sensitivity 16.3\% (subject-bootstrap interval 13.6--19.1\%), precision 14.0\%, and 0.230 unmatched alerts per decision-exposure hour. Median lead time among captured episodes was 6 minutes. The limited sensitivity and precision do not support a claim of clinically reliable early warning.

\begin{figure}[tbp]\centering\includegraphics[width=\linewidth]{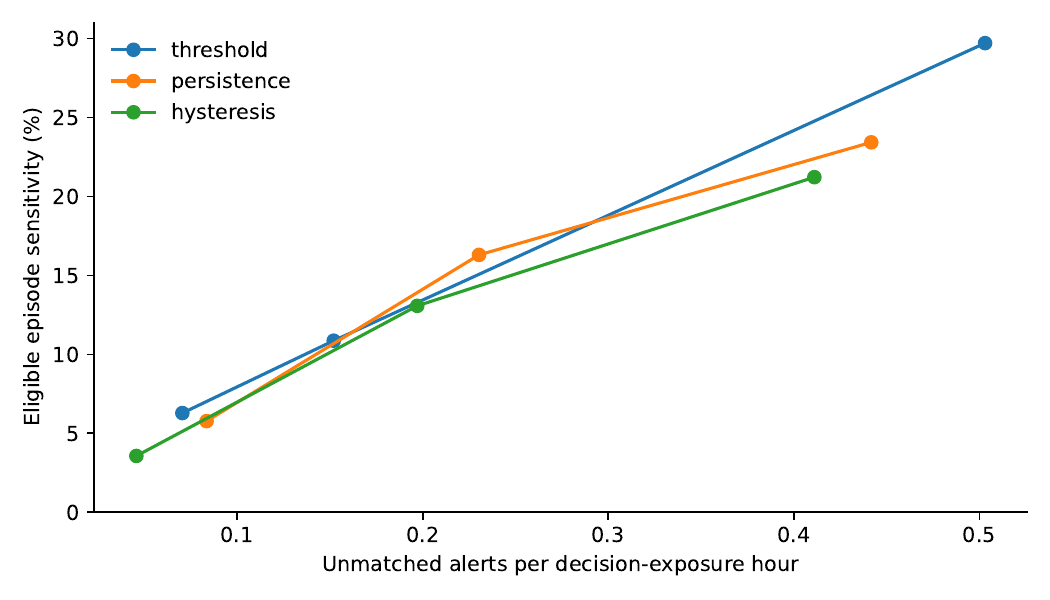}\caption{Test sensitivity and achieved unmatched-alert rate for validation-selected policies using direct prediction with additional MAP history. Points correspond to the three validation budgets; connecting lines aid identification and do not imply a continuous estimated frontier.}\end{figure}

\section{Discussion}
This benchmark evaluates future cumulative pressure deficit as a forecasting target and separates its statistical accuracy from alert behaviour. The direct and residual formulations share data, features and tree settings, which makes their comparison more informative than a comparison between unrelated architectures. The last-MAP-only model tests how much of the performance can be obtained from the current pressure. Additional MAP-history features probe whether depth and duration already present in the observed signal provide useful information beyond summary minute-pressure features.

The development iteration illustrates why implementation details must be reported. Multiplying all sample weights by a constant preserves their relative patient weighting, but changes their interaction with regularisation and minimum child weight. Consequently, apparently equivalent descriptions of ``patient-weighted XGBoost'' need not imply comparable fitted models. We report both conventions on validation and retain the mean-one convention for the frozen evaluation. The earlier handoff's numbers are not pooled with these results.

Calibration deserves explicit evaluation rather than inference from another performance measure, a recurring concern in clinical predictive analytics \cite{vancalster2019}. A well-calibrated upper quantile may support a graded display, but aggregate coverage is not a guarantee for an individual patient or subgroup. More generally, prediction-model evaluation distinguishes statistical performance from clinical usefulness \cite{steyerberg2010}. The burden distribution has a point mass at zero, so strict and non-strict coverage can differ substantially, particularly at the median. Separate quantile heads can cross; the current models should not be presented as a coherent full predictive distribution without further work. Monotone rearrangement is an established way to address crossing \cite{chernozhukov2010}, but it was not introduced after inspection of the frozen test results. Forecast loss improvements alone do not establish that clinicians would act differently or that their actions would improve outcomes.

Activation-based matching also changes the apparent utility of a warning rule. Crediting every episode that occurs while a persistent alert remains active can reward an uninformative always-on policy. Our one-to-one rule prevents that failure. Conversely, a persistent alert may remain clinically relevant; the present metric intentionally measures new early warnings, not continuous bedside awareness. Prospective workflow design should distinguish these functions rather than optimising an ambiguous alert count. Alarm fatigue is an established patient-safety concern \cite{sendelbach2013}. Observational intensive-care monitoring studies document substantial alarm burden \cite{drew2014}. Yoon and colleagues evaluated hypotension warning with a lockout rule in intensive care \cite{yoon2020}, while Hyland and colleagues evaluated a circulatory-failure warning system with an explicit silencing policy \cite{hyland2020}. Their event definitions and time horizons differ from ours; we use these studies to motivate explicit policy reporting, not to compare sensitivity figures across incompatible tasks.

Decision-curve analysis links prediction thresholds to clinical consequences through net benefit \cite{vickers2006}. Our unmatched-alert budgets express an operational constraint and do not constitute a net-benefit analysis: we have not specified the utility of an intervention or the relative clinical harms of missed and unnecessary warnings. Workflow evaluation should also measure response time, given observed associations between nonactionable alarm exposure and slower responses in hospital monitoring \cite{bonafide2015}. Similarly, external evaluation of a widely implemented sepsis model illustrates the importance of testing a deployed prediction system outside its development setting \cite{wong2021}; it is supporting evidence for the validation requirement, not evidence about our hypotension model.

\subsection{Machine learning, LLMs and pretrained forecasting models}
The evaluated machine-learning approach maps recent physiological summaries to conditional quantiles of future burden. Direct boosting learns this mapping from the observed target, whereas residual boosting learns a correction to a persistence forecast. These matched formulations test whether the representation of the learning problem adds value beyond current pressure. The test results support learned burden forecasting relative to persistence, but do not establish that greater model complexity is beneficial: the added MAP-history features did not demonstrate a clear incremental gain, and improved forecast loss did not translate into adequate episode-warning sensitivity under the evaluated alert budget.

A multichannel extension could explore multi-view graph representations \cite{xing2024multi}, but would require a justified graph construction and a matched comparison against the current feature-based models. Structured spatial reasoning in other modalities offers additional context for representation design \cite{yu2025core3d}.

Progress-aligned recurrence offers a separate methodological perspective on length generalisation in sequential state tracking \cite{wang2026parprogressalignedrecurrencelength}. Joint spatial-temporal forecasting in computer networks offers a separate methodological analogy for modelling dependencies across channels and time \cite{xing2025netsight}. Its relevance to physiological forecasting would require direct evaluation on clinical signals.

Application-specific control objectives have also been explored outside healthcare \cite{xing2025reinforcement}. For clinical decision support, action selection would require explicit treatment actions, clinically justified objectives and evaluation of intervention effects. Progress-aware robotic policy learning illustrates a further, nonclinical setting in which action selection explicitly tracks task progression \cite{liu2026palm}. Our fixed alert policies do not learn treatment actions, and the retrospective burden results do not establish the safety or effectiveness of a reinforcement-learning controller.

LLMs and pretrained time-series models offer possible extensions, but their roles should be distinguished. A model that produces clinical text is not, by that capability alone, a calibrated numerical forecaster. Chronos provides existing context for pretrained time-series forecasting \cite{ansari2024chronos}, while DT-GPT provides context for clinical trajectory modelling \cite{makarov2025}. Neither was evaluated here. A future comparison should preserve the same available history, subject-disjoint evaluation, burden definition and alert policies, and report pretraining-data provenance, adaptation requirements and inference cost. Efficiency-aware architectural comparisons have also been explored in state-space diffusion models for molecular generation \cite{das2025statespace}, although their performance does not establish suitability for physiological forecasting. If a candidate predicts pressure trajectories rather than burden directly, the comparison must propagate uncertainty across the full future trajectory to the accumulated deficit; integrating a single predicted path does not establish a conditional burden quantile.

An LLM could also be investigated as an interface for summarising model outputs or explaining the evaluation to clinicians. Taxonomy-aware adaptation provides context for specialising medical language models across clinical tasks \cite{dong2026conditionalrankallocationtaxonomyaware}. Language-guided policy specification has been explored in public-health allocation \cite{behari2024dlm}; that decision-support role is distinct from numerical burden prediction. If visual inputs were added, work on latent visual reasoning would provide methodological context \cite{li2026toward}. That would be a separate informatics task requiring assessment of factual consistency with the numerical forecast, communication of uncertainty and effects on clinical decisions. For visual explanations, spatial grounding would warrant separate evaluation, as reflected in work on fine-grained preference optimisation for spatial reasoning \cite{shen2026fine}. Such an interface was neither implemented nor tested in this study, and generated explanations should not be treated as evidence of forecast validity or treatment benefit. Codex's assistance in developing code and drafting this manuscript is disclosed separately; it was not a patient-level forecasting model in the benchmark.

A literature-supported interface would require content-grounded retrieval \cite{ma-etal-2026-intragent}, evaluated separately from the accuracy of the physiological forecast. Generated explanations should remain traceable to their supporting sources. Biomedical conclusion generation from structured evidence is a distinct evaluation task \cite{li2026medconclusion}.

Contribution-aware retention of historical evidence has been explored in long-horizon manufacturing inspection \cite{liu2026conmemcontributionawarememorylonghorizon}. Persistent agent memory would require access-control testing across interactions \cite{rao2026fragfusebypassingaccesscontrol}. No memory-enabled agent interface was implemented in this study. Broader proposals for foundation-model-assisted social-impact workflows likewise emphasise human oversight across system development and evaluation \cite{zhao2025social}.

Responses to knowledge-intensive scientific requests would also require dedicated safety evaluation \cite{jiang2025sosbench}. Forecast accuracy alone cannot establish safe behaviour for an accompanying conversational system.

Detection and attribution of generated text address provenance rather than factual correctness or clinical usefulness \cite{rao2026twobirds}. Research on human-perceived video fakeness similarly treats perceived artefacts as a specific evaluation target \cite{fu2025learning}. These distinct evaluation tasks do not validate the current forecasting models.

\subsection{Limitations and next validation requirements}
This is an internal test within one hospital's selected arterial-line surgical population. It does not establish transportability to other centres, non-invasive measurement or intensive care. Invasive and concomitant cuff measurements can differ, so changing the measurement modality requires explicit evaluation \cite{wax2011}. Relative clocks prevent calendar-time validation. Generalisation to changing populations is also an explicit research objective in pretrained sequential allocation models \cite{zhao2024pretrained}; it remains untested for our forecasting pipeline. Complete future-label requirements and retrospective anaesthesia boundaries create selection effects. Excluding incomplete observations can introduce bias when missingness is related to the underlying clinical process \cite{sterne2009}. Alert replay relaxes future completeness but still conditions on recorded boundaries and sufficient history; incomplete recordings can affect event availability and classification.

Numeric range screening does not identify all artefacts, and the jump filter is not a validated cleaning algorithm. Underdamping and resonance can distort arterial-pressure measurements even when values remain physiologically plausible \cite{romagnoli2014}. Systematic data errors can also affect downstream sequential allocation policies \cite{zhao2025whisperer}. Clinical outcomes, treatment responses and prospective data latency were not evaluated. Observed future pressure reflects existing treatment. A warning backtest therefore cannot identify untreated counterfactual trajectories or the benefit of responding to a new alert. Subgroup analyses are descriptive and do not establish fairness. Fairness in downstream resource allocation requires explicit objectives beyond subgroup forecast scores \cite{verma2024fairness}. No external cohort, calibrated pressure-path comparator or foundation-model experiment was completed. External validation should assess the unchanged model in a relevant independent population and evaluate both predictive performance and usefulness \cite{riley2024}. Resources such as MIMIC-IV \cite{johnson2023} provide a distinct clinical data setting; any external study would first need to verify arterial-signal availability, temporal resolution and endpoint compatibility rather than assume an electronic health record dataset supplies interchangeable signals.

This study is a retrospective internal evaluation. Observational reporting guidance supports transparent description of design, participant selection and limitations \cite{vonelm2007}. Its primary test evaluation is frozen, so future methodological changes require a new evaluation design or external cohort rather than repeated tuning to these test results. This separation of development and validation is consistent with established clinical prediction-model workflows \cite{steyerberg2014}. Reporting follows the principles of transparent prediction-model evaluation described in TRIPOD+AI \cite{collins2024tripod}. A formal risk-of-bias and applicability assessment would provide an additional appraisal of participant selection, predictors, outcomes and analysis \cite{wolff2019}; no completed PROBAST assessment is claimed here. A subsequent live clinical evaluation should use the stage-appropriate DECIDE-AI guidance \cite{vasey2022}; an eventual randomised intervention trial would instead require trial reporting that accounts for the AI component, as in CONSORT-AI \cite{liu2020}. Neither standard converts the present retrospective study into prospective evidence.

\section{Conclusion}
Conditional burden quantiles provide an evaluable alternative to binary hypotension forecasts. Patient-disjoint scoring, explicit weight scaling and activation-based alert matching are necessary to interpret the resulting performance. The completed internal benchmark supports further investigation while leaving external validity, clinical workflow impact and patient benefit unresolved.

\section*{Declarations}
\textbf{Funding:} This research received no funding.

\textbf{Competing interests:} The author declares no competing interests.

\textbf{Ethics and consent:} This noncommercial study used publicly available, de-identified VitalDB data. The source data collection and public release were approved by the Institutional Review Board of Seoul National University Hospital (H-1408-101-605), with written informed consent waived because the data were anonymised \cite{lee2022vitaldb,lee2022physionet}. No participants were recruited and no intervention was performed in this secondary analysis.

\textbf{Data and code availability:} VitalDB v1.0.0 is available at \url{https://doi.org/10.13026/czw8-9p62} under CC BY 4.0; numeric tracks were retrieved through the official API. Analysis code, pinned requirements, aggregate results and the freeze manifest accompany this manuscript.

\textbf{Author contributions:} Yunfan Zhao led the research ideas and conceptualisation, experimental design, project direction, manuscript writing, and revision and polishing of the paper.

\textbf{Declaration of generative AI assistance:} OpenAI Codex assisted with code reconstruction, experiment execution and manuscript drafting under the author's direction. The author led the intellectual development, experimental design and writing of the paper and takes responsibility for its content.

\end{document}